\documentclass{article}

\usepackage{microtype}
\usepackage{graphicx}
\usepackage{subfigure}
\usepackage{booktabs} 

\usepackage{hyperref}

\usepackage[accepted]{icml2021}

\icmltitlerunning{Trade-offs in Utility, Fairness and Differential Privacy}

\begin{document}

\twocolumn[
\icmltitle{Investigating Trade-offs in Utility, Fairness and Differential Privacy in Neural Networks}



\icmlsetsymbol{equal}{*}

\begin{icmlauthorlist}
\icmlauthor{Marlotte Pannekoek}{to}
\icmlauthor{Giacomo Spigler}{to}
\end{icmlauthorlist}

\icmlaffiliation{to}{Department of Cognitive Science and Artificial Intelligence, Tilburg University, Tilburg, Netherlands}

\icmlcorrespondingauthor{Giacomo Spigler}{g.spigler@uvt.nl}

\icmlkeywords{Machine Learning, fairness, differential privacy, utility-fairness-privacy tradeoff, neural networks}

\vskip 0.3in
]



\printAffiliationsAndNotice{}  

\begin{abstract}

To enable an ethical and legal use of machine learning algorithms, they must both be fair and protect the privacy of those whose data are being used. However, implementing privacy and fairness constraints might come at the cost of utility \cite{jayaraman2019evaluating, gong_preserving_2020}. This paper investigates the privacy-utility-fairness trade-off in neural networks by comparing a Simple (S-NN), a Fair (F-NN), a Differentially Private (DP-NN), and a Differentially Private and Fair Neural Network (DPF-NN) to evaluate differences in performance on metrics for privacy ($\epsilon$, $\delta$), fairness (risk difference), and utility (accuracy). In the scenario with the highest considered privacy guarantees ($\epsilon$ = 0.1, $\delta$ = 0.00001), the DPF-NN was found to achieve better risk difference than all the other neural networks with only a marginally lower accuracy than the S-NN and DP-NN. This model is considered fair as it achieved a risk difference below the strict (0.05) and lenient (0.1) thresholds. However, while the accuracy of the proposed model improved on previous work from Xu, Yuan and Wu \yrcite{xu2019achieving}, the risk difference was found to be worse.

\end{abstract}

\section{Introduction}
\label{introduction}

Machine learning algorithms are employed with great goals in mind, such as improved decision-making or increased efficiency. These benefits explain the ubiquitous use of such algorithms. However, potential harms should not be overlooked. Firstly, using personal data for training can cause leakage of private information. In addition, individuals can be unfairly affected when outcomes are dependent on race, sex, religious beliefs, sexual orientation, or economic status. For machine learning algorithms trained on personal data to be viable in today's society, they must ensure fairness and privacy for their users. This is required for both ethical and legal reasons \cite{xu2019achieving, wood_differential_2018}. In this paper, we build on existing research to investigate how algorithms can ensure privacy while remaining fair and useful. To this extent, we tested how applying the fairness method Reject Option Classification influences the performance of a neural network \cite{briggs2020mitigating}. Furthermore, the effects of adding a privacy-preserving optimizer to a simple and to a fair neural network were investigated. 

Prior research has shown that when personal data are used for training machine learning models, these data can be retrieved by observing the behavior or structure of the learning algorithms \cite{yeom_overfitting_2020}. This can negatively affect social status, employment chances, and insurance costs for end users of the systems \cite{wood_differential_2018}. Therefore, it appears critical to develop and apply privacy-preserving methods. Along this direction, here we focus on the differential privacy framework, which is currently considered state-of-the-art \cite{gong_preserving_2020}. Its goal is to produce an approximately equal output, whether an individual is included in the analysis or not \cite{wood_differential_2018}. An essential element of differential privacy is the privacy budget $\epsilon$, which controls how well privacy is protected. In general, lower values of $\epsilon$ imply more privacy protection. Nonetheless, choosing the best value for the parameter is still difficult \cite{jayaraman2019evaluating}.

Finally, fairness  entails that the behavior of a machine learning system does not dependent on protected attributes such as race and gender \cite{zhu2020more}. It is often assumed that algorithms are fair and impartial since decisions are based on data instead of human judgment \cite{corbett-davies_measure_2018}. However, several studies provide evidence that algorithms can be negatively biased towards protected groups, which could, for example, cause men to be preferred over women with similar skills by hiring algorithms \cite{xu2019achieving}. 
Several methods have been developed to achieve fairness via manipulation of the data or of the algorithms during pre-processing, in-processing, or post-processing phases. In this research, the post-processing method of ``Reject Option Classification'' was applied to a simple and a differentially private neural network to explore its effect on the performance of the network.

The existing empirical research on differentially private and fair algorithms has mainly focused on logistic regression \cite{xu2019achieving, jagielski_differentially_nodate, ding_differentially_2020}. Research by Friedler et al.  \yrcite{friedler_comparative_2019}, however, shows that the trade-off between accuracy and fairness can differ greatly depending on the specific algorithm being used. This suggests that it would be useful to research the combination of fairness and privacy on a wider range of machine learning systems.

In recent years both differential privacy and fairness have received much attention \cite{caton_fairness_2020, yeom_overfitting_2020}. However, several authors have argued more research on algorithms that can achieve both goals \cite{ekstrand_privacy_nodate, zhu2020more, xu2019achieving, datta_correspondences_2018}.  Earlier work on differential privacy has shown that there can be trade-offs between utility and privacy-preservation \cite{jayaraman2019evaluating, gong_preserving_2020}. Similar work has been done on the trade-off between utility and fairness \cite{feldman_certifying_2015}. \emph{Here we investigate the trade-off between utility, privacy, and fairness when the goal of an algorithm is to perform well on all three metrics. }

Specifically, we explore how the privacy-utility-fairness trade-off in neural networks is affected by:\\
- the Reject Option Classification method for fairness;\\
- a differentially private optimizer, under varying privacy budgets;\\
- the joint use of a differentially private optimizer and the Reject Option Classification method for fairness.

We finally compare the different cases with corresponding state-of-the-art results by Xu, Yuan, and Wu \yrcite{xu2019achieving}.\\

\section{Related Work}

The combination of fairness and differential privacy has been the subject of recent empirical research \cite{xu2019achieving}. The work by Xu, Yuan and Wu \yrcite{xu2019achieving} in particular compared two algorithms with respect to utility and fairness, under different choices of the privacy budget ($\epsilon$). The algorithms were tested on two benchmarks; the Adult and the Dutch dataset. The metrics used for utility and fairness were accuracy and risk difference, respectively. For their first algorithm, logistic regression was used, and a penalty was added to the objective function to ensure fairness. A technique known as the functional mechanism was then applied to the objective function to ensure differential privacy. In the second algorithm, the objective function was corrupted by noise sampled from a Laplace distribution of which the mean was shifted according to a fairness constraint, therefore satisfying both fairness and differential privacy. Both algorithms achieved differential privacy and fairness with reasonable utility. Their results show that as $\epsilon$ was decreased, thus increasing the degree of privacy protection, the accuracy decreased, with a dynamics consistent across both datasets and methods.


Further, Jagielski et al. \yrcite{jagielski_differentially_nodate} compared the use of a post-processing method for achieving fairness with an in-processing method. Only the algorithm using post-processing achieved a reasonable fairness-accuracy-privacy trade-off. However, only privacy with regard to the sensitive attribute was taken into account.

Ding et al. \yrcite{ding_differentially_2020} also combined privacy and fairness and explored varying the amount of noise added to different attributes. They also used a less strict definition of differential privacy. Using these techniques, they improved on the results achieved by Xu, Yuan and Wu \yrcite{xu2019achieving}. Their results suggest that increasing $\epsilon$ from 0.1 to 1 results in a higher increase in accuracy than increasing 
$\epsilon$ from 10 to 100. However, this difference was not tested for significance. For most values of $\epsilon$, their algorithms achieved reasonable values of risk difference. 

Cummings, Gupta, Kimpara and Morgenstern \yrcite{cummings_compatibility_2019} found that exact fairness and differential privacy in their set-up could not be achieved, although a trade-off could be achieved by relaxing the notion of exact fairness to `approximate fairness'.

Hajian, Domingo-Ferrer, Monreale, Pedreschi and Giannotti \yrcite{hajian_discrimination-_2015} investigated the application of privacy-preserving and fairness techniques on patterns derived by a classifier using multiple association rules. Their approach can be classified as a post-processing approach as they altered the classifier's results instead of the classifier itself or the data it was trained on \cite{hajian_discrimination-_2015}. They investigate both \emph{k}-anonymity and differential privacy to achieve privacy. Their empirical analyses demonstrate that \emph{k}-anonymity distorted the patterns less than the differential privacy approach. Additionally, their results show that applying the privacy technique after the fairness technique can deteriorate the achieved fairness.

\section{Experiments}\label{experiments}


\subsection{Dataset Description}\label{3.1Dataset}
The experiments presented here were performed using the Adult dataset \cite{Dua:2019}, due to its importance in research on privacy and fairness, and to allow for a direct comparison with state-of-the-art results \cite{xu2019achieving}.

The importance of the Adult dataset in the field is due to the presence of sensitive attributes that could be used for personal identification or potentially threaten the fairness of models trained on it. The dataset is openly available and consists of 45,222 cases and 14 variables. In this research, only 'sex' was regarded as a sensitive attribute, and 'income' was regarded as the dependent variable. Income is treated as a binary variable, separating incomes of less than 50,000 (income = 0) and more than 50,000 (income = 1). The sex variable refers to biological sex with male and female as the possible values. 

\subsection{Data Pre-Processing}\label{3.2Datapro}
To ensure comparability, the same pre-processing steps were applied as in the study by Xu, Yuan and Wu \yrcite{xu2019achieving}. Specifically, list-wise deletions were performed, dummy codes were used for the categorical variables, and continuous variables were normalized. The Adult dataset already contained a separate train and test set. The train dataset was further split up into train and validation sets. The train, validation, and test dataset constitute 53.4\%, 13.3\%, and 33.3\% of the total amount of data, respectively. 

Initial data exploration showed a class imbalance in the labels, with approximately 75$\%$ of the reported incomes being less than 50,000. Furthermore, data exploration showed that females were underrepresented, making up around 32$\%$ of the cases. Approximately 88$\%$ of females earned less than 50,000, against 69$\%$ of males, which motivated applying fairness constraints.

\subsection{Models}\label{3.4models}
Four models were compared to explore the effect of differential privacy and fairness methods on the privacy-utility-fairness trade-off: a baseline `Simple' neural network (S-NN), and a Fair (F-NN), a Differentially Private (DP-NN), and a Differentially Private and Fair neural network (DPF-NN). All models were implemented using Keras \cite{chollet2015keras}. When no specific parameter settings are mentioned, the default settings were used. 

\textbf{Simple Neural Network (S-NN).}
The S-NN consisted of three fully connected layers with six neurons in the first and second layer and one neuron in the final layer. The first and the second layer used a ReLu activation, while the last layer used a sigmoid activation. Binary cross-entropy was used as the loss function and Adam as the optimizer \cite{kingma2017adam}. Training was performed for a fixed duration of 20 epochs of Stochastic Gradient Descent with minibatches of size $mb=20$.

\textbf{Fair Neural Network (F-NN).}
The network used for the F-NN was equal to the S-NN. However, the `Reject Option Classification' method was added to alter the output labels after prediction in an effort to improve fairness. The method was chosen as the best performing of six fairness methods that were previously evaluated. Results from the comparison are reported in the Supplementary Materials. The pre-processing and post-processing techniques to ensure fairness were implemented using the Artificial Intelligence Fairness 360 library (AIF 360) \cite{aif360-oct-2018}.


\textbf{Differentially Private Neural Network (DP-NN).}
The network used for the DP-NN was equal to the S-NN. However, training was performed using a differentially private variant of the Adam optimizer (DPAdamGaussianOptimizer) \cite{mcmahan_general_2019}. This optimizer adds Gaussian noise to the gradient to ensure differential privacy. The noise$\_$multiplier parameter was used to specify the amount of noise added to the model. This parameter's value depends on the target value for $\epsilon$ and $\delta$ (quantifying the probability of not achieving privacy within the privacy budget.), and was calculated using the compute$\_$dp$\_$sgd$\_$privacy function. Another notable aspect of the differentially private optimizer is that norm clipping is applied after the data are split up into minibatches, but before adding the noise. Training was repeated for values of $\epsilon \in \{0.1, 1, 10, 100\}$ and $\delta \in \{0.01, 0.001, 0.0001, 0.00001\}$, as they are commonly used values in differential privacy applications \cite{xu2019achieving, ding_differentially_2020}.

\textbf{Differentially Private and Fair Neural Network (DPF-NN).}

The DPF-NN finally integrated the F-NN and the DP-NN, combining the use of the Reject Option Classification method from F-NN with the differentially private optimizer from DP-NN. The same values of $\epsilon$ and $\delta$ as the DP-NN were used.

\subsection{Evaluation Criteria}\label{3.6Evaluation}
The different models were tested for accuracy and risk difference to assess the utility and fairness of the models, respectively. All experiments were repeated ten times with different random seeds. This procedure is compatible with the previous work by Xu, Yuan and Wu \yrcite{xu2019achieving}. 

A mean accuracy above 75.4$\%$ on test data was considered an improvement over an algorithm that chooses the majority label. Risk difference was used as the metric to evaluate model fairness. A risk difference of 0 was considered optimal for fairness. Standard thresholds below which models are considered fair include 0.05 and 0.1 \cite{briggs2020mitigating, xu2019achieving}. Both were considered and will be referred to as the strict and the lenient threshold, respectively, throughout this paper.  

The risk difference and accuracy scores (for given values of $\epsilon$ and $\delta$) were compared to the thresholds and the scores achieved by the baseline models by Xu, Yuan and Wu \yrcite{xu2019achieving}. Independent t-tests were applied to assess significant differences between algorithms, using a significance level of 0.05. Lastly, linear regression was performed to test for an effect of $\epsilon$ and $\delta$ on mean accuracy in the DP-NN and DPF-NN models.

\section{Results} \label{Results}

The accuracy and risk difference for all the neural networks ($\epsilon = 0.1$, $\delta = 0.00001$) are shown in Table \ref{tab:comparison}, together with the results achieved by the equivalent models from Xu, Yuan and Wu \yrcite{xu2019achieving}. Further comparisons between the models are available in the Supplementary Materials. 


\begin{table}[htb!]
\caption{Comparison between the performance of the neural networks from this work (top half) and the logistic regression models by Xu, Yuan and Wu \yrcite{xu2019achieving} (bottom half). The order of the models indicate correspondence between the models proposed in this work and those of Xu, Yuan and Wu (simple baseline - LR; fair network - FairLR; differentially private network - PrivLR; differentially private and fair network - PFLR*). The privacy parameters $\epsilon = 0.1$, $\delta = 0.00001$ were used in the models that applied differential privacy constraints. Performance is shown as mean accuracy (in percentage) and risk difference (with standard deviations). Higher values are preferred for accuracy, whereas lower values are preferred for risk difference.}
\label{tab:comparison}
\vskip 0.15in
\begin{center}
\begin{small}
\begin{sc}
\begin{tabular}{lrr}
\toprule
            & Accuracy             & Risk difference     \\ \midrule
S-NN         & 84.14 ± 0.34      & 0.1310 ± 0.0147     \\
DP-NN        & 84.03 ± 0.05      & 0.1355 ± 0.0024     \\
F-NN         & 79.25 ± 3.50      & 0.0566 ± 0.0065     \\
DPF-NN       & 82.98 ± 0.19      & 0.0475 ± 0.0020     \\ \hline
LR          & 83.80 ± 0.23      & 0.1577 ± 0.0064     \\
PrivLR      & 62.63 ± 14.80      & 0.0883 ± 0.0805     \\
FairLR      & 77.39 ± 5.21      & 0.0095 ± 0.0071     \\
PFLR*       & 74.91 ± 0.40      & 0.0028 ± 0.0039      \\
\bottomrule
\end{tabular}
\end{sc}
\end{small}
\vskip 0.15in
\end{center}

\end{table}

T-tests were applied to assess whether the differences between the different neural networks and the models by Xu, Yuan and Wu \yrcite{xu2019achieving} are significant at a significance level of 0.05. The results from these t-tests regarding the mean accuracy and risk difference scores can be found in Table \ref{tab:ttestaccuracy} and Table \ref{tab:ttestrd}, respectively.



\begin{table*}[h]
\tabcolsep=0.15cm
\caption{Difference in mean accuracy between all models. A positive score means that the model defined in the row performed better than the model defined in the column. Eq. Model refers to the equivalent model from Xu, Yuan and Wu \yrcite{xu2019achieving}. For example, FairLR is the equivalent model for the F-NN. An independent t-test was performed to test if the difference in means was significant at significance level $\alpha = .05$ (indicated by *). $DF = 18$ for all t-tests. The t-statistic is reported in brackets. The privacy parameters $\epsilon = 0.1$, $\delta = 0.00001$ were used in the models that applied differential privacy constraints. }
\label{tab:ttestaccuracy}
\vskip 0.15in
\begin{center}
\begin{small}
\begin{sc}
\begin{tabular}{lrrrrr}
\toprule
      & S-NN            & DP-NN           & F-NN            & DPF-NN           & Eq. Model \\ \midrule
S-NN   & 0 (0.0)        & 0.11 (1.0)   & 4.89* (4.4) & 1.16* (9.4)  & 0.34* (2.6) \\
DP-NN  &                & 0 (0.0)       & 4.78* (4.3) & 1.05* (16.9)  & 21.40* (4.6) \\
F-NN   &                &               & 0 (0.0)     & -3.73* (-3.4)  & 1.86 (0.9) \\
DPF-NN &  &  & & 0 (0.0)         & 8.07* (57.6)\\ \bottomrule
\end{tabular}
\vskip 0.15in
\end{sc}
\end{small}
\end{center}
\end{table*}


\begin{table*}[]
\tabcolsep=0.12cm
\caption{Difference in mean risk difference between all models. A negative score means that the model defined in the row performed better than the model defined in the column. Eq. Model refers to the equivalent model from Xu, Yuan and Wu \yrcite{xu2019achieving}. For example, FairLR is the equivalent model for the F-NN.* means that the difference is significant at a 0.05 significance level. $DF = 18$ for all t-tests. The t-statistic is reported in brackets. The privacy parameters $\epsilon = 0.1$, $\delta = 0.00001$ were used in the models that applied differential privacy constraints. }
\label{tab:ttestrd}
\vskip 0.15in
\begin{center}
\begin{small}
\begin{sc}
\begin{tabular}{lrrrrr}
\toprule
      & S-NN             & DP-NN              & F-NN           & DPF-NN         & Eq. Model \\ \midrule
S-NN   & 0 (0.0)         & -0.0045 (-1.0)    & 0.0744* (14.6)& 0.0835* (17.8)& -0.0267* (-5.3)  \\
DP-NN  &                 & 0  (0.0)          & 0.0789* (36.0)& 0.0880* (89.1)& 0.0472 (1.9)   \\
F-NN   &                 &                   & 0 (0.0)       & 0.0091* (4.2)& 0.0471* (15.5) \\
DPF-NN &                  &                   &             & 0  (0.0)      & 0.0447* (32.3)  \\ \bottomrule
\end{tabular}
\vskip 0.15in
\end{sc}
\end{small}
\end{center}
\end{table*}

\textbf{Simple Neural Network (S-NN).}\label{4.1S-NN}
As shown in Table \ref{tab:comparison} the S-NN achieved an average accuracy of 84.14\% (\emph{SD} = 0.34). This is slightly but significantly higher than the simple logistic regression (LR) by Xu, Yuan and Wu \yrcite{xu2019achieving}, \emph{t}(18) = 2.6, \emph{p} = .017, which achieved an average accuracy of 83.80\% (\emph{SD} = 0.23). The average risk difference for the S-NN was 0.1310 (\emph{SD} = 0.0147), which is slightly but significantly lower than the 0.1577 (\emph{SD} = 0.00064) risk difference from the model by Xu, Yuan and Wu \yrcite{xu2019achieving}, \emph{t}(18) = -5.3, \emph{p} $<$ .001. The achieved accuracy of the S-NN is above the threshold of an algorithm that chooses the majority label. 

\textbf{Fair Neural Network (F-NN).}\label{4.2F-NN}



The F-NN achieved an average accuracy of 79.25\% (\emph{SD} =  3.50), which is above the majority label threshold and a significant decrease of 4.89 compared to the S-NN (\emph{M} = 84.14\%, \emph{SD} = 0.34), \emph{t}(18) = 4.4, \emph{p} $<$ .001. Compared to the average accuracy from the fair model by Xu, Yuan and Wu \yrcite{xu2019achieving} (\emph{M} = 77.39\%, \emph{SD} = 5.21), this is an increase of 0.0186 in average accuracy. However, this difference is not significant, \emph{t}(18) = 0.9, \emph{p} = .361. The mean risk difference achieved by the F-NN is 0.0566 (\emph{SD} = 0.0065), which is slightly above the 0.05 threshold. Compared to the risk difference of the S-NN (\emph{M} = 0.1310, \emph{SD} = 0.0147), this is a 0.0744 decrease, which was found to be significant, \emph{t}(18) = -14.6, \emph{p} $<$ .001. Compared to the fair logistic regression model by Xu, Yuan and Wu \yrcite{xu2019achieving}, which achieved an average risk difference of 0.0095 (\emph{SD} = 0.0071), this is a significant increase of 0.0471, \emph{t}(18) = 15.5, \emph{p} $<$ .001.

In conclusion, the application of Reject Option Classification in the F-NN did lead to a decreased risk difference compared to the S-NN. The mean risk of the F-NN was below the lenient 0.1 threshold. However, it is still slightly above the 0.05 threshold and higher than the average risk difference achieved by the fair logistic regression model by Xu, Yuan and Wu \yrcite{xu2019achieving}. The fair model's accuracy is lower compared to the S-NN but still acceptable and higher than that of the fair logistic model. 

\textbf{Differentially Private Neural Network (DP-NN).}\label{4.3DP-NN}

A table can be found in the appendix that displays the mean accuracy and risk difference for the DP-NN with differing values for $\epsilon$ and $\delta$. A linear regression was run to determine whether there was a significant effect of $\epsilon$ and $\delta$ on average accuracy and risk difference for the DP-NN. No significant effect on risk difference ($F(6,9) = 3.15, p = .0597, R^2 = 0.68$) or accuracy ($F(6,9) = 2.88, p = .0748, R^2 = 0.66$) could be observed for varying values of $\epsilon$ and $\delta$. Across all $\delta$ and $\epsilon$ values, the overall average of the average accuracy is 84.05\% (\emph{SD} = 0.05). The overall average of the average risk difference is 0.1345 (\emph{SD} = 0.0016).

The average accuracy of the DP-NN model with $\epsilon$ = 0.1 and $\delta$ = 0.00001, so with the highest privacy guarantee, is 84.03\% (\emph{SD} = 0.05). With the highest privacy guarantee, the DP-NN achieved a mean accuracy that was 0.11 lower than that of the S-NN (\emph{M} = 84.14\%, \emph{SD} = 0.34). This difference was, however, not significant (\emph{t}(18) = -1.0, \emph{p} = .325). The average risk difference of the model with the highest privacy guarantee is 0.1355 (\emph{SD} = 0.0024), a difference of 0.0045 compared to the S-NN (\emph{M} = 0.1310, \emph{SD} = 0.0147). However, this difference is not significant, \emph{t}(18) = 1.0, \emph{p} = .352.

In the appendix a summary table is given of the accuracy and risk difference for differing values of $\epsilon$ and $\delta$ = 0.00001 for the DP-NN and the differentially private logistic regression model by Xu, Yuan and Wu \yrcite{xu2019achieving}. When comparing the models with the lowest $\epsilon$, the DP-NN model's average risk difference (\emph{M} = 0.1355, \emph{SD} = 0.0024) is 0.0472 higher than that of the differentially private logistic regression (\emph{M} = 0.0883, \emph{SD} = 0.0805). However, this is not a significant difference, \emph{t}(18) = 1.9, \emph{p} = .080. The DP-NN model (\emph{M} = 84.03\%, \emph{SD} = 0.05) does significantly improve the average accuracy by 21.40, \emph{t}(18) = 4.6, \emph{p} $<$ .001, in comparison with the differentially private logistic regression (\emph{M} = 62.63\%, \emph{SD} = 14.80). When comparing the models with the highest $\epsilon$, DP-NN also achieved a higher average accuracy (\emph{M} = 84.03\%, \emph{SD} = 0.05) in comparison with the logistic regression model (\emph{M} = 82.95\%, \emph{SD} = 0.32). The difference is smaller (10.08) but still significant, \emph{t}(18) = 10.5, \emph{p} $<$ .001. 

In conclusion, no trend in average accuracy or risk difference was found for varying values of $\epsilon$ and $\delta$. Furthermore, there were no significant differences in average accuracy or risk difference between the DP-NN and the S-NN ($\epsilon$ = 0.1, $\delta$ = 0.00001). The DP-NN does, however, improve the average accuracy compared to the differentially private logistic regression.

\textbf{Differentially Private and Fair Neural Network (DPF-NN).}\label{4.4DPF-NN}
In the appendix a table is provided that shows the average accuracy and risk difference for varying values of $\epsilon$ and $\delta$ for the DPF-NN. As with the DP-NN, the mean risk difference and the accuracy barely differ for different values of $\delta$ and $\epsilon$. This is supported by the results from a simple linear regression that was performed to determine whether there was a significant effect of $\epsilon$ and $\delta$ on mean accuracy and risk difference. These results show no significant effect on mean accuracy ($F(6,9)= 0.66, p = .687, R^2 = 0.30$) or risk difference ($F(6,9) = 0.57, p = .748, R^2 = 0.27$). 
Across all values of $\delta$ and $\epsilon$ the average of the mean accuracy scores is 82.96 \% (\emph{SD} = 0.25). The overall average for the average risk difference is 0.0475 (\emph{SD} = 0.0017). All averages for risk difference, including the overall average, are below the 0.05 threshold. 

The DPF-NN with the highest privacy guarantee ($\epsilon$ = 0.1 and $\delta$ = 0.00001) has a mean accuracy of 82.98\% (\emph{SD} = 0.19). Compared to the S-NN (\emph{M} = 84.14\%, \emph{SD} = 0.34), that is a 1.16 lower average accuracy. This is a significant difference (\emph{t}(18) = -9.4, \emph{p} $<$ .001).
The DPF-NN with the highest privacy guarantee achieved a mean risk difference of 0.0475 (\emph{SD} = 0.0020), which is a 0.0835 lower average risk difference compared to the S-NN (\emph{M} = 0.1310, \emph{SD} = 0.0147). This is also a significant difference, \emph{t}(18) = -17.8, \emph{p} $<$ .001.  

Compared to the F-NN (\emph{M} = 79.25\%, \emph{SD} = 3.50), the most private DPF-NN (\emph{M} = 82.98\%, \emph{SD} = 0.19) has a significantly higher average accuracy (\emph{t}(18) = 3.4, \emph{p} = .003). The difference between the models is 3.73. The average risk difference of the DPF-NN (\emph{M} = 0.0475, \emph{SD} = 0.0020) is 0.0091 lower compared to the most private F-NN (\emph{M} = 0.0566, \emph{SD} = 0.0065), which is a significant difference (\emph{t}(18) = -4.2, \emph{p} = .001).

When $\epsilon$ equals 0.1 and $\delta$ equals 0.00001, the average accuracy of the DPF-NN (\emph{M} = 82.98\%, \emph{SD} = 0.19) is significantly lower by 1.05 (\emph{t}(18) = -16.9, \emph{p} $<$ .001) compared to the DP-NN (\emph{M} = 84.03\%, \emph{SD} = 0.05). With respect to average risk difference, the DPF-NN (\emph{M} = 0.0475, \emph{SD} = 0.0020) has a significantly lower average compared to the DP-NN (\emph{M} = 0.1355, \emph{SD} = 0.0024). This difference of 0.0880 is significant (\emph{t}(18) = -89.1, \emph{p} $<$ .001).  
 
When the results for $\delta$ = 0.00001 for the DPF-NN are compared to the results from the PFLR* model by Xu, Yuan and Wu \yrcite{xu2019achieving}, some conclusions can be drawn. Firstly, for all values of $\epsilon$, the PFLR* model achieves lower average risk difference than the DPF-NN model. However, the average accuracy scores are also lower for the PFLR*. The difference in risk difference between the two models is at most 0.0447, with the lowest $\epsilon$. In this case the difference between the DPF-NN (\emph{M} = 0.0475, \emph{SD} = 0.0020) and the PFLR* (\emph{M} = 0.0028, \emph{SD} = 0.0039) is significant (\emph{t}(18) = 32.3, \emph{p} $<$ .001).
The minimal difference in risk difference between the DPF-NN (\emph{M} = 0.0437, \emph{SD} = 0.0154) and the PFLR* (\emph{M} = 0.0204, \emph{SD} = 0.0140) is 0.0233 when $\epsilon$ equals 10. This difference is still significant (\emph{t}(18) = 3.5, \emph{p} = .002). 

For average accuracy, the largest difference between the DPF-NN (\emph{M} = 82.98\%, \emph{SD} = 0.19) and the PFLR* (\emph{M} = 74.91\%, \emph{SD} = 0.40) was 8.07, for the lowest privacy budget. This is a significant difference (\emph{t}(18) = 57.6, \emph{p} $<$ .001).
The smallest difference in average accuracy between the DPF-NN (\emph{M} = 83.04\%, \emph{SD} = 0.23) and the PFLR* (\emph{M} = 79.13\%, \emph{SD} = 2.00) was for the highest $\epsilon$ and came down to a difference of 3.91. This smallest difference is also significant (\emph{t}(18) = 6.1, \emph{p} $<$ .001).

In conclusion, adding differential privacy and fairness to a simple neural network decreased the average risk difference below the 0.05 threshold. The average accuracy also decreased but was still well above the 75.4\% threshold. Adding both differential privacy and fairness compared to only adding fairness increased the accuracy and decreased the risk difference. Adding both differential privacy and fairness compared to only adding differential privacy decreased the risk difference and slightly decreased the accuracy. When the DPF-NN ($\delta$ = 0.00001) is compared to the DPFLR* by Xu, Yuan and Wu \yrcite{xu2019achieving}, the DPF-NN achieves a higher average risk difference, though still under the strict 0.05 threshold, but higher average accuracy for all values of $\epsilon$.

\section{Discussion}

Four models were compared, a Simple (S-NN), a Fair (F-NN), a Differentially Private (DP-NN), and a Differentially Private and Fair Neural Network (DPF-NN). These models were evaluated for different values of $\epsilon$ and $\delta$ on fairness and utility metrics. The models were compared relative to each other, to the models by Xu, Yuand and Wu \yrcite{xu2019achieving}, and to several threshold values.



\textbf{Effects of Fairness Constraints on the Fairness-Utility Trade-off.}

Adding fairness constraints to the S-NN significantly reduced the mean accuracy but also significantly decreased the mean risk difference. The F-NN model achieved a mean risk difference of 0.0566 (\emph{SD} = 0.0065). This is below the lenient 0.1 threshold. However, it is close to but not below the strict 0.05 threshold. The achieved accuracy was above the threshold of choosing the majority label. Reductions in statistical parity also decreased utility when Reject Option Classification was applied in the research by Hufthammer, et al. \yrcite {hufthammer2020bias}. However, in the research by Briggs and Hollmén \yrcite{briggs2020mitigating}, the utility was improved. 

\textbf{Effects of Privacy Constraints on the Privacy-Utility Trade-off.}
The performance of the DP-NN was evaluated for different values of $\epsilon$ and $\delta$. A linear regression showed that no significant effect on risk differences or accuracy could be observed as a function of $\epsilon$ and $\delta$. Furthermore, changing the standard Adam optimizer of the S-NN to the differentially private optimizer in the DP-NN did not significantly change the accuracy nor the risk difference when the highest privacy constraints were applied ($\epsilon$ = 0.1, $\delta$ = 0.00001). This model achieved an accuracy well above the majority label baseline. The risk difference was above both the strict and the lenient thresholds, which is expected since no fairness constraints were added to this model. \emph{Most notably and contrary to previous results \cite{zhao_not_2020, jayaraman2019evaluating, gong_preserving_2020}, no disruption in accuracy was observed when using lower values of $\epsilon$ and $\delta$.}

\textbf{Effects of Fairness and Privacy Constraints on the Privacy-Utility-Fairness Trade-off.}
Like the case of DP-NN, no trend in risk differences or accuracy was observed for varying values of $\epsilon$ and $\delta$ in the DPF-NN model. The results on the DPF-NN showed that adding both fairness and a strong privacy guarantee ($\epsilon$ = 0.1, $\delta$ = 0.00001) to the S-NN model significantly increased its accuracy. However, this also significantly decreased risk difference. While the accuracy of the DPF-NN with $\epsilon$ = 0.1 and $\delta$ = 0.00001 was lower, it was still well above the majority label baseline. The achieved mean risk difference by this model was below both the strict and lenient thresholds. The model is, therefore, considered fair. Adding both fairness and differential privacy ($\epsilon$ = 0.1, $\delta$ = 0.00001) significantly increased accuracy compared to only adding fairness, but it decreased accuracy compared to only adding differential privacy ($\epsilon$ = 0.1, $\delta$ = 0.00001). Adding both differential privacy and fairness improved fairness compared to adding only differential privacy or fairness.

\textbf{Comparison with Equivalent Models.}
The results from Xu, yuan and Wu \yrcite{xu2019achieving} were used as baselines to compare the performance of the proposed models.

The S-NN achieved a significantly higher mean accuracy and a lower mean risk difference than the baseline simple model. 
The F-NN achieved higher accuracy than the baseline fair model though this difference is not significant. The F-NN also produced a significantly higher mean risk difference and is, therefore, deemed less fair.
There were no significant changes in risk difference between the DP-NN with the highest privacy guarantees ($\epsilon$ = 0.1,  $\delta$ = 0.00001) and the PFLR* with the highest privacy guarantees ($\epsilon$ = 0.1) by Xu, Yuan and Wu \yrcite{xu2019achieving}. The DP-NN did, however, achieve a significantly higher mean accuracy. 
The DPF-NN with the highest privacy guarantees ($\epsilon$ = 0.1,  $\delta$ = 0.00001) significantly outperformed the baseline differentially private and fair model with the highest privacy guarantees ($\epsilon$ = 0.1) on mean accuracy but produced a significantly higher mean risk difference. It is, thus, less fair but has more utility.

\textbf{Limitations and Future Research.}

In the present work, only the variable `sex' was considered as a sensitive attribute. However, in many practical applications, multiple sensitive variables may need to be considered. For example, the models may be considered fair with respect to gender but still make discriminatory decisions with respect to race. Likewise, Caton and Haas \yrcite{caton_fairness_2020} also warned of the effects of variables that are themselves not considered sensitive, but are still related to sensitive variables. Furthermore, only one fairness metric was considered in this research, risk difference, which is a measure of demographic/statistical parity that captures group fairness. As previously discussed, reduced unfairness according to one metric may not reduce and it may even increase unfairness according to another metric \cite{lee_evaluation_2020, hufthammer2020bias, caton_fairness_2020}. If the models used in this research would be employed in real-life settings it would, therefore, be crucial to consider if demographic/statistical parity would be suitable for the specific application. Related to this are the chosen threshold: for example, in this research, the limit of risk difference beneath which a model was deemed fair was 0.05 for the strict threshold and 0.1 for the lenient threshold. However, whether these thresholds are suitable in a real-life application may depend on the use case and legislative requirements \cite{datta_correspondences_2018}. If the goal of fairness is preferred over utility, the fair and differentially private and fair logistic regression models by Xu, Yuan and Wu \yrcite{xu2019achieving} should be preferred over the equivalent models presented here. Likewise, utility may be measured differently in different applications, which may require using balanced accuracy, F1-score, or other metrics, which might lead to different model rankings. Additionally, it would be interesting to assess whether combining pre- and post- processing fairness techniques would improve results. Lastly, a relaxed notion of differential privacy was considered in this research, while in some applications, strict or traditional differential privacy may be preferred. 


Lastly, the results achieved by the differentially private and fair model in this research are encouraging and should be validated on a larger selection of datasets and using other metrics to assess the privacy-utility-fairness trade-off.

\section{Conclusion}

In this paper, we explored the impact of differential privacy and fairness constraints on the privacy-utility-fairness, both when they are applied independently and when they are combined together. Contrary to the previous research on this topic, this research focused on neural networks instead of logistic regression. Applying only fairness constraints led to a model with high accuracy and fairness but no privacy. Applying only privacy constraints led to a private but unfair model with high accuracy. The model that combined privacy and fairness constraints achieved better fairness than the model that only applied fairness constraints, while maintaining high accuracy. While the accuracy of the fair and private model significantly improved on previous work from Xu, Yuan and Wu \yrcite{xu2019achieving}, the risk difference was found to be worse. Contrary to previous research both the DP-NN and DPF-NN model did not show a trend of a decrease in accuracy with an increase in the offered privacy. 
In conclusion, creating models that achieve fairness and preserve privacy while maintaining satisfactory utility is both possible and necessary. Hopefully, this and other contributions to the existing research on private and fair models can encourage and improve their use in real-world applications.








\nocite{langley00}

\bibliography{example_paper}
\bibliographystyle{icml2021}


\end{document}